\documentclass[sigconf]{acmart}

\usepackage{xspace}
\usepackage{algorithm}
\usepackage{algpseudocode}
\usepackage{amsmath}
\usepackage{booktabs}   
\usepackage{multirow}   
\usepackage{graphicx}   
\usepackage{pifont}     
\usepackage{xcolor}
\usepackage{balance}
\newcommand{\cmark}{\textcolor{green!60!black}{\ding{51}}}
\newcommand{\xmark}{\textcolor{red!70!black}{\ding{55}}}

\newcommand{\mstd}[2]{#1 {\scriptsize $\pm$ #2}}

\newcommand{\TReNN}{\textsc{TReNN}\xspace}
\newcommand{\ReNN}{\textsc{ReNN}\xspace}
\newcommand{\TNN}{\textsc{TNN}\xspace}
\newcommand{\SNN}{\textsc{SNN}\xspace}
\newcommand{\TEReNN}{\textsc{TE-ReNN}\xspace}
\AtBeginDocument{%
  }

\setcopyright{acmlicensed}\copyrightyear{2026}
\acmYear{2026}
\setcopyright{cc}
\setcctype{by}
\acmConference[MM '26]{Proceedings of the 34th ACM International Conference on Multimedia}{November 10--14, 2026}{Rio de Janeiro, Brazil}
\acmBooktitle{Proceedings of the 34th ACM International Conference on Multimedia (MM '26), November 10--14, 2026, Rio de Janeiro, Brazil}
\acmDOI{10.1145/3767308.3836430}
\acmISBN{979-8-4007-2213-4/2026/11}

\begin{document}

\title[Counterfactual Relational Modeling of Surgical Teams]{
  From Multimodal Observation to Interpretable Suggestions:
  Counterfactual Time-Expanded Relational Modeling of Surgical Teams
}
 \author{Vincenzo Marco De Luca}
 \email{vincenzomarco.deluca@unitn.it}
 \orcid{0009-0003-3209-5817}
 \affiliation{%
   \institution{University of Trento}
   \city{Trento}
   \country{Italy}
 }

 \author{Antonio Longa}
 \email{antonio.longa@uit.no}
 \orcid{0000-0003-0337-1838}
 \affiliation{%
   \institution{UiT the Arctic University of Norway}
   \city{Tromsø}
   \country{Norway}
 }

 \author{Giovanna Varni}
 \email{giovanna.varni@unitn.it}
  \orcid{0000-0002-5659-8414}

 \affiliation{%
   \institution{University of Trento}
   \city{Trento}
   \country{Italy}}

 \author{Andrea Passerini}
 \email{andrea.passerini@unitn.it}
 \orcid{0000-0002-2765-5395}
 \affiliation{%
   \institution{University of Trento}
   \city{Trento}
   \country{Italy}
 }

\begin{abstract}
In surgery, patient safety is threatened not only by technical issues but also by poor teamwork. However, existing surgical AI-based solutions focus mainly on visual workflow and technical execution, neglecting the modeling of team interactions and missing opportunities to actively support clinicians in improving their teamwork skills.
To address this gap, we propose a tempo-relational framework for modeling surgical team dynamics from multimodal observations. By leveraging Time-Expanded graphs, the approach captures both relational structure and temporal evolution, achieving strong expressivity while remaining robust in the low-data regime typical of surgical settings.
Beyond prediction, 
such modeling enables the generation of efficient, interpretable, and actionable suggestions for clinicians. More specifically, we generate suggestions via a counterfactual procedure that identifies minimal yet structured changes in individual behaviors and interaction patterns associated with improvements in team performance. 
Experiments with simulated surgical procedures show that our approach improves predictive performance in diverse behavioral and interaction goals while offering meaningful insights into team dynamics. This work advances surgical AI beyond outcome-driven prediction towards a socially grounded, team-centric, and actionable paradigm to better understand and support the development of team skills in surgical settings.
\end{abstract}



\maketitle

\section{Introduction}
Actionable AI aims to move beyond prediction by enabling systems that provide interpretable and actionable feedback to support humans in their work and everyday tasks. This requirement is critical in high-stakes settings, such as operating rooms (ORs), in which team performance arises from complex, dynamic, and tightly coordinated interactions. Accurate predictions alone are insufficient; decision support must translate the system's outputs into actionable guidance that enhances team performance.
The OR scenario exemplifies such high-stakes settings. While Surgical Data Science (SDS) has made significant progress in modeling procedural workflows~\cite{vedula2017surgical, maier2017surgical, ward2021surgical, kawka2022intraoperative} and individual technical skills~\cite{lam2022machine, kostopoulos2025prediction}, modeling intra-operative team dynamics and providing actionable insights remain largely underexplored. 
Building computational models that capture temporal-relational patterns in small-team interactions is particularly challenging in surgical settings due to limited data availability and its complex technical setup. Additionally, high-dimensional multimodal data require generalizable models that can leverage all modalities without overfitting.

In this work, we propose a graph-based multimodal team modeling approach particularly suitable for high-stakes scenarios. More specifically, in our setting, the model predicts social and behavioral constructs representing the overall team dynamics. To capture temporal evolution directly within the relational structure, we construct Time-Expanded graphs, enabling message passing along both spatial and temporal dimensions within a single architecture. This approach preserves the expressive power of the model while remaining robust in low-data regimes.
Building on this team representation, we introduce a counterfactual approach 
for actionable reasoning. Unlike previous approaches, which focus solely on the removal of edges~\cite{lucic2022cf} or combinatorial enumeration of possible alternative behaviors~\cite{qu2025cody}, our method identifies minimal and interpretable team behavioral changes to improve teamwork outcomes based on the previous history observed in the training data. 
These counterfactual insights are validated in surgical training settings, in which clinical teams perform procedures on high-fidelity simulators, ensuring both validity and reproducibility.

The contributions of this paper are as follows:
\begin{itemize}
\item We propose \TEReNN, a Time-Expanded Relational Neural Network that models multimodal team interactions 
while remaining robust in the low-data regime; 
\item We introduce a graph-based counterfactual approach that provides interpretable and actionable recommendations to improve teamwork;
\item We demonstrate that our approach outperforms existing approaches in surgical training simulation, highlighting robust multimodal modeling, team-level prediction, and validated counterfactual insights.
\end{itemize}

\section{Related Work}

Automated team modeling has attracted growing interest, but progress remains limited by the difficulty of collecting multimodal data in multiparty settings and the scarcity of large-scale datasets~\cite{kraaij2005ami,sanchez2011audio,beyan2016detecting,braley2018group,bhattacharya2019unobtrusive,muller2021multimediate}. Early work focus on static team modeling, in which multimodal features are aggregated across individuals and used to predict team constructs such as cohesion, leadership, or performance via classical ML or shallow neural models~\cite{hung-gatica-perez-2010,nanninga-2017,beyan2017moving,murray2018predicting,kubasova2019analyzing,muller2024multimediate}. However, these approaches ignore temporal dynamics and interpersonal interactions. Temporal models address this limitation by capturing behavioral evolution over time using sequence-based methods such as HMMs, LSTMs, and Transformers~\cite{avci2016predicting,lin2023interaction,10191640}, as well as dynamical systems and imitation learning~\cite{gorman2017understanding,seo2023automated}. While effective at modeling temporal dependencies, they typically treat team members independently and overlook relational structure. Conversely, relational approaches incorporate interactions among team members through graph-based or network-based models, including Graph Neural Networks (GNNs) and attention mechanisms~\cite{lin2020predicting,sharma2023graphitti,bai2023dips,li2024dat}, but often neglect temporal dynamics or focus on limited tasks. Recent temporal GNNs attempt to bridge this gap~\cite{longa2023graph}, yet they are rarely applied to multi-construct team modeling and often lack comprehensive evaluation. Overall, existing approaches either model time or relations in isolation, leaving a gap for unified approaches that jointly capture temporal, relational, and multimodal aspects of team behavior. A notable exception is the recent work by De Luca et al.~\cite{Luca2026}, which introduces tempo-relational neural networks to model teamwork. However, their approach suffers from the low-data regime typical of medical team analysis, where models are trained on a set of recordings. Indeed, the approach has only been evaluated on much simpler cooperative survival games performed in the laboratory. Our approach sidesteps this problem by directly encoding temporal dependencies within the graph structure. This design enables effective message-passing while remaining robust in the low-data and small team regime typical of operating room scenarios. Our experimental evaluation demonstrates that this constitutes a substantial advantage over alternative approaches.
Human-AI collaboration research has demonstrated that hybrid teams outperform humans or AI alone only when interaction is well-calibrated and systems support transparent and adaptive cooperation~\cite{andrews2023role, mateos2025spatio}. Trust calibration and shared mental models are key determinants of team performance that can be supported by multiple explanation modalities while mitigating over-reliance on AI recommendations~\cite{herlocker2000explaining, eiband2018bringing}. In particular, counterfactual explanations provide a formal framework to estimate the effects of hypothetical interventions, enabling systems to reason about what could have happened under alternative actions. For this reason, they have been widely adopted in machine learning to interpret model decisions, often as post-hoc diagnostic tools applied to individual predictions~\cite{wachter2017counterfactual}. In high-stakes domains such as healthcare, counterfactual methods have been used to guide treatment selection~\cite{johansson2016learning, schwab2020learning}, yet these approaches rarely consider interactions among multiple decision makers or provide structured feedback that can inform team-level decision processes. Similarly, counterfactual policy learning methods optimize intervention outcomes but often ignore the temporal and social dynamics inherent in collaborative environments.
Some recent studies integrate temporal and relational modeling to support counterfactual generation~\cite{qu2025cody, lutraining}, yet these methods are often used as predictive or explanatory tools rather than mechanisms for actionable feedback. From a computational perspective, small-team scenarios introduce additional challenges due to limited data availability, sparsity, and the absence of publicly labeled datasets.
Despite these advances, existing approaches present notable limitations. Most counterfactual methods neglect the complex interactions among multiple agents over time. Explanations are rarely translated into actionable guidance for workflow decisions, and the dynamic evolution of teams during collaboration is often unmodeled. Our work addresses these gaps with Time-Expanded Relational Neural Networks (\TEReNN), which generate structured counterfactual alternatives capturing both temporal evolution and social interaction patterns. Unlike prior methods, \TEReNN provides actionable feedback at the level of individual decisions, team members' interactions, and collective outcomes, supporting team decision-making in high-stakes operational contexts rather than producing post-hoc explanations detached from the decision process.

\begin{figure*}[t!]
    \centering
    \includegraphics[width=0.9\textwidth]{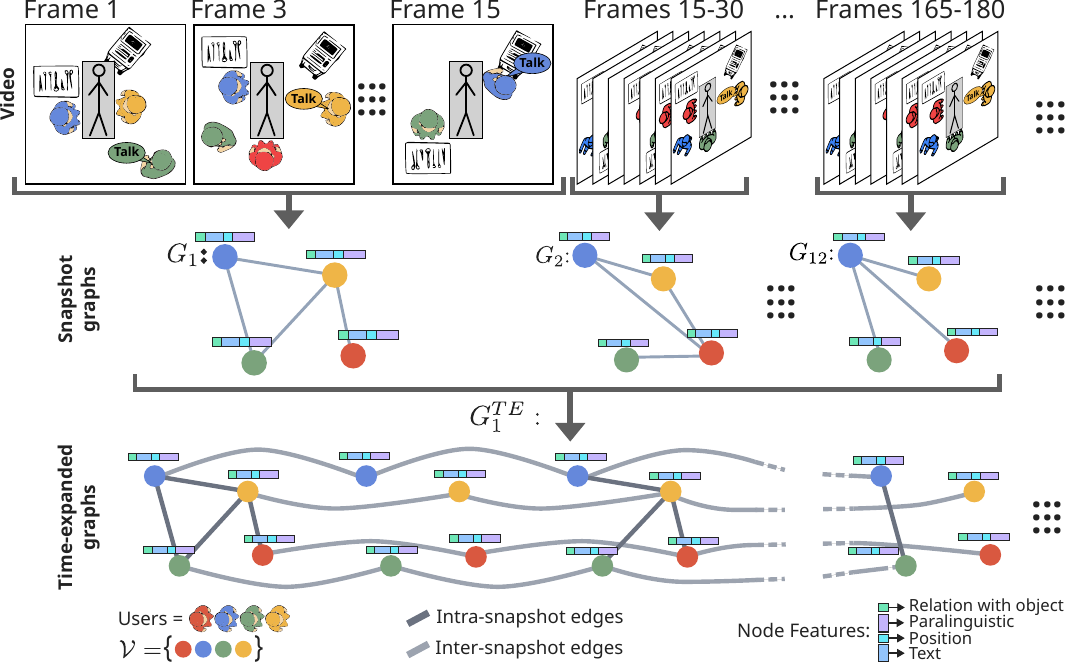}
    \caption{Overview of the interaction modeling pipeline. 
    Top row: multimodal time-series are segmented into fixed temporal windows (15 seconds). 
    Middle row: for each window, a snapshot interaction graph $G_t$ is built, in which nodes represent team members and edges encode broadcast verbal communication. Node features integrate interpretable paralinguistic (eGeMAPS), text, position, and human--tool interaction (relation with objects). 
    Bottom row: snapshot graphs are connected through identity-preserving temporal edges into a Time-Expanded Graph $G^{TE}$, enabling joint modeling of team members' coordination and longitudinal individual dynamics across time.}
    \label{fig:main_panel}
\end{figure*}

\section{Method}
\subsection{Time-Expanded Relational Neural Networks}

We represent each surgical team as a sequence of multimodal interaction graphs built over fixed temporal windows. The input comprises (i) multimodal features extracted from synchronized audio-video recordings and annotations, including acoustic, motion, and human--object interaction cues, and (ii) team conversation transcripts. Since these modalities are acquired at different sampling rates, all features were aggregated into aligned non-overlapping windows of 15 seconds.

Let $\mathcal{V}$ denote the set of all team members involved in the surgery. For each temporal window $t \in \{1, \dots, T\}$, we first build a \emph{snapshot interaction graph}
\[
G_t = (V_t, E_t),
\]
where $V_t \subseteq \mathcal{V}$ is the set of team members present during window $t$. Edges in $E_t$ encode verbal communication assuming that speech is broadcast to all team members present: whenever a team member $v_i \in V_t$ speaks during window $t$, directed edges $(v_i, v_j)$ are added from the speaker $v_i$ to all the other team members present in that window $v_j \in V_t$ with $j \neq i$. This assumption models verbal communication as a one-to-many interaction, reflecting the shared acoustic environment of the OR.
Each node $v \in V_t$ is associated with a multimodal feature vector composed of:
\begin{itemize}
    \item \textbf{Paralinguistic features}: loudness, alpha-ratio, and harmonics-to-noise ratio from the eGeMAPS 2.0 set~\cite{eyben2015geneva}. The selection of this subset is detailed in Sec.~\ref{sec:preprocessing}.
    \item \textbf{Text features}: language-model embeddings extracted from the speech transcription associated with each team member speaking within a temporal window.
    \item \textbf{Motion features}: the spatial position $(x,y)$ of each team member, together with the mean and standard deviation of their displacement within a temporal window.
    \item \textbf{Human--tool interaction features}: the triplet \emph{team member---action---object} describing atomic actions in the OR (e.g., an\-es\-the\-si\-ol\-o\-gist---calibrating---instrument). They are encoded via one-hot vectors and associated with the corresponding team member.
    \item \textbf{Role features}: a categorical label reporting the functional role of each team member (e.g., head surgeon, nurse), encoded using a one-hot representation.
\end{itemize}

To capture team coordination dynamics beyond the window-based resolution, we transform the sequence of snapshot graphs into a \emph{Time-Expanded graph}, which explicitly encodes temporal evolution in the graph topology, as illustrated in Fig.~\ref{fig:main_panel}. Given a temporal sequence of snapshots
\[
\mathcal{G}_{1:T} = [G_1, \dots, G_T],
\]
we define the corresponding Time-Expanded graph as
\[
G^{TE} = (V^{TE}, E^{TE}),
\]
with
\[
V^{TE} = \{ v_i^t \mid v_i \in V_t,\; t = 1, \dots, T \}.
\]
Each node $v_i^t$ represents team member $v_i$ at the temporal window $t$. Consequently, multiple temporal instances of the same team member coexist in the graph, allowing their behavior to be tracked throughout the surgical procedure.
To capture both instantaneous and longitudinal dependencies, the edge set is defined as
\[
E^{TE} = E^{snap} \cup E^{temp},
\]
where:
\begin{itemize}
    \item \textbf{Intra-snapshot edges}:
    \[
    E^{snap} = \{ (v_i^t, v_j^t) \mid (v_i, v_j) \in E_t \}.
    \]
    These edges preserve the communication structure observed within each temporal window and capture broadcast interactions from a speaking team member $v_i^t$ to the other team members $v_j^t$ in the same snapshot.

    \item \textbf{Inter-snapshot identity edges}:
    \[
    E^{temp} = \{ (v_i^t, v_i^{t+1}) \mid v_i \in V_t \cap V_{t+1},\; t = 1, \dots, T-1 \}.
    \]
    These edges connect consecutive temporal instances of team members and preserve identity continuity over time.
\end{itemize}

This construction encodes temporal dynamics directly into the graph topology, allowing information to propagate both across team members within each window and along the temporal trajectory of every individual.
In our implementation, each Time-Expanded graph spans up to six minutes of audio-video recordings, that is corresponding to 24 temporal windows, providing sufficient temporal context to capture evolving coordination patterns while remaining computationally tractable. 
By combining relational structure with interpretable behavioral abstraction, the proposed representation enables the learning of predictive signals of procedural efficiency while maintaining a direct link between model representations and human-adjustable behaviors.
This representation is naturally compatible with Message Passing Graph Neural Networks (MP-GNNs)~\cite{micheli2009neural,kipf2017semi,pmlr-v70-gilmer17a,Wu_2021}, as it defines a unified graph space in which information can be iteratively aggregated from both relational and temporal neighborhoods. In particular, the intra-snapshot edges enable the exchange of contextual information across team members within the same temporal window, while the inter-snapshot identity edges enable each representation to evolve along its own temporal trajectory. This design extends standard message passing to a tempo-relational setting, where node embeddings are updated by jointly integrating social interactions and their temporal evolution, without requiring explicit recurrent mechanisms \cite{gao2022equivalence}. The resulting Time-Expanded Relational Neural Network (\TEReNN) is better suited to low-data settings than recurrent-based approaches, as it encodes temporal dependencies directly within the graph structure rather than relying on sequential state propagation, thus better leveraging the message-passing capabilities of graph neural network architectures. In the low-data regime typical of surgical procedures, in which recordings are scarce and annotations are costly, this leads to consistently improved predictive performance across tasks, as demonstrated in our experimental evaluation.

\subsection{Actionable Counterfactuals}
\label{sec:actionable_counterfactual}
We generate actionable feedback by identifying how a team interaction sequence could evolve toward a more desirable outcome. Let
\[
\mathcal{G}_{1:T} = [G_1, \dots, G_T]
\]
denote an input sequence of snapshot graphs, and let $y^*$ denote a target outcome (e.g., high cooperation). The goal is to extract interpretable modifications to team behaviors that steer the prediction of the model towards $y^*$.
Note that in this setting, in which multiple team members interact over extended time horizons, standard counterfactual approaches that seek minimal changes to flip a model’s prediction~\cite{wachter2017counterfactual,moth2020} are not directly applicable. Modifying the behavior of a single team member (e.g., the lead surgeon) at a given time step may yield a more favorable predicted outcome, but such a counterfactual is unrealistic, as it ignores the cascading effects on other team members and their behaviors in subsequent time steps.
To sidestep this issue, we adopt an exemplar-based strategy grounded in real observations, following example-based counterfactual literature~\cite{10.1007/978-3-030-86957-1_3, brughmans2024nice}, where explanations are obtained by comparing an instance with a similar but more desirable example. Accordingly, throughout the paper we refer to our method as an example-based counterfactual explanation. Given a sequence exhibiting suboptimal team performance, we retrieve a sequence with a similar temporal prefix but a desirable outcome, and compare the two prefixes to identify the key behavioral differences underlying the divergence in performance.

\paragraph{Overview.}
The method consists of two stages: (i) offline construction of an embedding repository, and (ii) online counterfactual generation through retrieval, explanation, and comparison.

\paragraph{Embedding repository.}
We precompute a repository $\mathcal{Z}$ of prefix-level embeddings from the training set, as detailed in Algorithm~\ref{alg:repository}. For each training sequence
\[
\mathcal{G}_{1:T_i}^i = [G_1^i, \dots, G_{T_i}^i],
\]
we consider all prefixes
\[
\mathcal{G}_{1:t}^i = [G_1^i, \dots, G_t^i], \qquad t = 1, \dots, T_i.
\]
Each prefix is converted into its corresponding Time-Expanded graph and encoded using the trained GNN $f_\theta$, producing a graph embedding $\mathbf{z}^i_{\leq t}$. Each entry in $\mathcal{Z}$ stores the tuple
\[
(\mathbf{z}^i_{\leq t}, y^i, \mathcal{G}_{1:t}^i),
\]
where $y^i$ is the outcome label associated with sequence $i$. This step is performed once and reused at inference time, enabling efficient retrieval of temporally aligned interaction patterns.

\begin{algorithm}[t]
\caption{Embedding Repository Construction}
\label{alg:repository}
\begin{algorithmic}[1]
\Require Training set $\mathcal{D}$, encoder $f_\theta$
\Ensure Repository $\mathcal{Z}$

\State $\mathcal{Z} \gets \emptyset$
\For{each sequence $\mathcal{G}_{1:T_i}^i = [G_1^i, \dots, G_{T_i}^i] \in \mathcal{D}$}
    \For{$t = 1$ to $T_i$}
        \State $\mathcal{G}_{1:t}^i \gets [G_1^i, \dots, G_t^i]$
        \State construct the corresponding Time-Expanded graph $G_{i,\leq t}^{TE}$
        \State $\mathbf{z}^i_{\leq t} \gets f_\theta(G_{i,\leq t}^{TE})$
        \State $\mathcal{Z} \gets \mathcal{Z} \cup \{(\mathbf{z}^i_{\leq t}, y^i, \mathcal{G}_{1:t}^i)\}$
    \EndFor
\EndFor
\end{algorithmic}
\end{algorithm}

\paragraph{ii) Counterfactual generation.}
At inference time, we analyze the input incrementally using Algorithm~\ref{alg:counterfactual}. For each prefix
\[
\mathcal{G}_{1:t} = [G_1, \dots, G_t],
\]
we construct the corresponding Time-Expanded graph $G_{\leq t}^{TE}$ and encode it with $f_\theta$ to obtain the embedding $\mathbf{z}_{\leq t}$. We then query the embedding repository $\mathcal{Z}$ via the function \textsc{RetrieveImproved} to retrieve the embedding which is closest to $\mathbf{z}_{\leq t}$ according to cosine similarity, has the same temporal length $t$ and exhibits the desired outcome $y^*$:

\[
(\mathbf{z}^*_{\leq t}, y^*, \mathcal{G}_{1:t}^*)
\]

Next, we apply \textsc{Explain} to the retrieved prefix $\mathcal{G}_{1:t}^*$ to identify which elements most strongly support its prediction. Since faithfulness assessments for GNN explanations are sensitive to the underlying architecture \cite{azzolin2025reconsiderin}, we use saliency scores to rank candidate behavioral components. Starting from feature-level saliency scores~\cite{simonyan2013deep}, we derive importance estimates at the node and edge levels, and aggregate them into interpretable behavioral components, such as individual actions or communication cues. This yields an ordered explanation of the retrieved example, ranked from the most to the least influential components.

We then compare these components with the current prefix $\mathcal{G}_{1:t}$ using \textsc{Match}, where we exploit the role labels available in the graph to compare components at role-consistent positions, e.g., surgeon-to-surgeon or nurse-to-nurse. The candidate explanations are examined in decreasing order of importance, and the first salient component that is absent from the input is returned as a counterfactual suggestion. If all salient components are already present, the algorithm proceeds by extending the prefix. Further details on \textsc{RetrieveImproved}, \textsc{Explain}, and \textsc{Match} are reported in the Supplementary Material. 

\begin{algorithm}[t]
\caption{Counterfactual Generation}
\label{alg:counterfactual}
\begin{algorithmic}[1]
\Require Input sequence $\mathcal{G}_{1:T} = [G_1, \dots, G_T]$, target outcome $y^*$, repository $\mathcal{Z}$, encoder $f_\theta$
\Ensure Counterfactual explanation $e$

\For{$t = 1$ to $T$}
    \State $\mathcal{G}_{1:t} \gets [G_1, \dots, G_t]$
    \State construct the corresponding Time-Expanded graph $G_{\leq t}^{TE}$
    \State $\mathbf{z}_{\leq t} \gets f_\theta(G_{\leq t}^{TE})$

    \State $(\mathbf{z}^*_{\leq t}, y^*, \mathcal{G}_{1:t}^*) \gets \textsc{RetrieveImproved}(\mathbf{z}_{\leq t}, \mathcal{Z}, y^*)$

    \State $\mathcal{E} \gets \textsc{Explain}(\mathcal{G}_{1:t}^*, f_\theta)$

    \For{each candidate explanation $e \in \mathcal{E}$}
        \If{$\textsc{Match}(e, \mathcal{G}_{1:t}) = \textbf{false}$}
            \State \Return $e$
        \EndIf
    \EndFor
\EndFor

\State \Return $\emptyset$
\end{algorithmic}
\end{algorithm}
The proposed approach combines example-based retrieval with explanation-based refinement. Retrieval ensures candidate behaviors are realistic by grounding them in observed interaction sequences, while explanations isolate the components responsible for the desirable outcome. Comparing these components with the current interaction sequence yields actionable suggestions and identifies the earliest point an intervention could be recommended.

\section{Experimental setting}

\subsection{Dataset}

The MM-OR~\cite{ozsoy2025mm} dataset is a multimodal corpus of 21 simulated knee replacement surgical procedures, including 14 full and 7 partial knee replacement surgeries. In our experiments, we use 15 procedures, nine of which are full knee replacements. Each procedure typically involves four to six team members, whose functional roles are explicitly annotated, and may include either expert surgeons or trainees. 
The surgeries are audio-video recorded using five cameras and a single environmental microphone. The dataset also includes annotations about the use of specific tools by team members. 
Behavioral annotations were performed manually using the validated Observational Teamwork Assessment for Surgery (OTAS)~\cite{hull2011observational} behavioral rating schema, which includes five behavioral dimensions (\emph{Communication}, \emph{Coordination}, \emph{Cooperation}, \emph{Leadership}, and \emph{Situational Awareness}) structured across 15 items on a 7-point Likert scale. Three annotators independently scored the videos at six-minute intervals; the resulting scores were averaged and rounded to the nearest value on the Likert scale.

\begin{table}
\centering
\small
\begin{tabular}{c c c l}
\hline
\textbf{Activation} & \textbf{Control} & \textbf{Dominance} & \textbf{Behavioral Class} \\
\hline
Low  & Low  & Low  & Withdrawn / Disengaged \\
Low  & Low  & High & Restrained Conflict \\
Low  & High & Low  & Calm--Cooperative \\
Low  & High & High & Quiet Authority \\
High & Low  & Low  & Agitated / Over-aroused \\
High & Low  & High & Dominant--Aggressive \\
High & High & Low  & Engaged--Cooperative \\
High & High & High & Calm Leader \\
\hline
\end{tabular}
\caption{Behavioral classes defined by combining three paralinguistic features extracted from the eGeMAPS set.}
\label{tab:behavioral_classes}
\end{table}

\subsection{Pre-processing}
\label{sec:preprocessing}

Automated diarization of audio recordings was performed using Pyannote~\cite{bredin2020pyannote}, and transcripts were generated with Whisper~\cite{radford2023robust}. Due to the use of a single environmental microphone, background noise, and frequent code-switching between English and German, these automated methods proved unreliable. Therefore, two annotators manually corrected both the diarization and the transcripts to ensure high-quality data.
\begin{table*}[h!]
    \centering
    \begin{tabular}{l l c c c c c c c c c}
    \toprule
        Paradigm & Model 
        & \multicolumn{3}{c}{Modeling Dimension}  & \multicolumn{1}{c}{Avg. F1-macro (\%)}
        & \multicolumn{5}{c}{F1-macro Measures (\%)} \\
        \cmidrule(lr){3-5}  \cmidrule(lr){6-6} \cmidrule(lr){7-11}
        & & Feature & Time &  Topology 
        & Teamwork & Comm & Coord & Coop & Lead & Sit. Aw \\
        \midrule
        \multirow{2}{*}{\SNN} 
        & MLP   & \cmark & \xmark & \xmark & \mstd{59.8}{2.2} & \mstd{60.1}{2.0} & \mstd{58.7}{2.3} & \mstd{60.3}{2.5} & \mstd{59.2}{1.8} & \mstd{60.7}{2.2} \\
        & RF    & \cmark & \xmark & \xmark & \mstd{60.2}{0.8} & \mstd{61.0}{0.9} & \mstd{59.5}{0.5} & \mstd{60.0}{0.6} & \mstd{60.3}{1.0} & \mstd{60.3}{1.2} \\
        \midrule
        \multirow{3}{*}{\TNN} 
        & LSTM  & \cmark & \cmark & \xmark & \mstd{62.3}{1.6} & \mstd{63.0}{2.4} & \mstd{61.8}{1.4} & \mstd{62.5}{1.8} & \mstd{62.0}{1.1} & \mstd{62.3}{1.1} \\
        & GRU   & \cmark & \cmark & \xmark & \mstd{61.8}{1.8} & \mstd{62.2}{2.1} & \mstd{61.5}{1.7} & \mstd{61.7}{2.3} & \mstd{61.8}{1.7} & \mstd{61.8}{1.2} \\
        & MHA   & \cmark & \cmark & \xmark & \mstd{63.8}{2.0} & \mstd{64.0}{2.1} & \mstd{63.5}{2.5} & \mstd{63.9}{2.8} & \mstd{63.6}{1.4} & \mstd{63.9}{1.4} \\
        \midrule
        \multirow{3}{*}{\ReNN} 
        & GCN   & \cmark & \xmark & \cmark & \mstd{61.4}{1.5} & \mstd{61.6}{1.4} & \mstd{61.2}{1.2} & \mstd{61.5}{1.5} & \mstd{61.3}{1.5} & \mstd{61.5}{1.8} \\
        & GAT   & \cmark & \xmark & \cmark & \mstd{62.1}{1.8} & \mstd{62.4}{1.9} & \mstd{61.8}{2.4} & \mstd{62.0}{2.1} & \mstd{62.0}{1.2} & \mstd{62.2}{1.5}  \\
        & GIN   & \cmark & \xmark & \cmark & \mstd{65.2}{1.5} & \mstd{65.0}{1.6} & \mstd{65.1}{1.4} & \mstd{65.3}{1.3} & \mstd{65.2}{1.8} & \mstd{65.4}{1.9} \\
        \midrule
\multirow{3}{*}{\shortstack[l]{
Space-then-Time\\[-1pt]\TReNN}}& GCN+MHA & \cmark & \cmark & \cmark & \mstd{64.0}{1.5} & \mstd{64.1}{1.7} & \mstd{63.8}{1.9} & \mstd{64.2}{1.4} & \mstd{63.9}{1.2} & \mstd{64.0}{1.4} \\
        & GAT+MHA & \cmark & \cmark & \cmark & \mstd{67.2}{1.7} & \mstd{67.5}{2.2} & \mstd{66.8}{1.8} & \mstd{67.3}{1.8} & \mstd{67.0}{1.0} & \mstd{67.2}{1.5} \\
        & GIN+MHA & \cmark & \cmark & \cmark & \mstd{63.2}{1.5} & \mstd{63.0}{1.8} & \mstd{63.1}{1.5} & \mstd{63.3}{1.5} & \mstd{63.2}{1.3} & \mstd{63.2}{1.4} \\
        \midrule
\multirow{3}{*}{\shortstack[l]{Time-then-Space\\[-1pt]\TReNN}}   
& MHA+GCN & \cmark & \cmark & \cmark & \mstd{65.2}{1.3} & \mstd{64.6}{0.9} & \mstd{66.0}{1.8} & \mstd{66.0}{1.1} & \mstd{64.4}{1.8} & \mstd{65.0}{0.9} \\
        & MHA+GAT & \cmark & \cmark & \cmark & \mstd{65.8}{1.5} & \mstd{66.4}{1.7} & \mstd{65.2}{1.3} & \mstd{65.8}{1.1} & \mstd{65.0}{2.0} & \mstd{66.6}{1.4} \\
        & MHA+GIN & \cmark & \cmark & \cmark & \mstd{64.2}{1.2} & \mstd{64.6}{1.2} & \mstd{63.8}{0.7} & \mstd{64.4}{1.6} & \mstd{64.3}{1.6} & \mstd{63.9}{0.9} \\
        \midrule
\multirow{3}{*}{\shortstack[l]{Standard\\[-1pt]\TReNN}}
& TGN       & \cmark & \cmark & \cmark & \mstd{65.0}{1.4} & \mstd{64.8}{1.5} & \mstd{65.2}{1.6} & \mstd{65.3}{1.3} & \mstd{64.9}{1.4} & \mstd{65.0}{1.5} \\
& DySAT     & \cmark & \cmark & \cmark & \mstd{65.6}{1.3} & \mstd{65.8}{1.4} & \mstd{65.4}{1.5} & \mstd{65.7}{1.3} & \mstd{65.5}{1.1} & \mstd{65.6}{1.2} \\
& EvolveGCN & \cmark & \cmark & \cmark & \mstd{66.0}{1.3} & \mstd{65.8}{1.4} & \mstd{66.1}{1.6} & \mstd{66.2}{1.3} & \mstd{65.9}{1.4} & \mstd{66.0}{1.2} \\
        \midrule
        \multirow{3}{*}{\TEReNN}  
        & TE-GCN & \cmark & \cmark & \cmark & \mstd{72.4}{1.7} & \mstd{72.0}{1.9} & \mstd{72.2}{1.9} & \mstd{72.5}{1.4} & \mstd{72.3}{1.6} & \mstd{72.5}{1.6} \\
        & TE-GAT & \cmark & \cmark & \cmark & \mstd{73.1}{1.7} & \mstd{73.0}{1.7} & \mstd{73.2}{1.9} & \mstd{73.3}{1.8} & \mstd{73.1}{1.1} & \mstd{73.0}{1.9} \\
        & \textbf{TE-GIN} & \cmark & \cmark & \cmark & \textbf{\mstd{74.2}{1.7}} & \mstd{74.0}{1.8} & \mstd{74.1}{1.6} & \mstd{74.3}{1.9} & \mstd{74.2}{1.2} & \mstd{74.4}{2.1} \\
        \bottomrule
    \end{tabular}
    \caption{Average \textbf{F1-macro (\%) and standard deviation (\%)} over ten seeds. 
    Teamwork average F1-macro denotes the mean across five OTAS dimensions: Communication (Comm), Coordination (Coord), Cooperation (Coop), Leadership (Lead), and Situational Awareness (Sit. Aw). 
    Bold indicates the best overall performance.}
    \label{tab:main_results}
\end{table*}
\begin{figure*}
    \centering
    \includegraphics[width=0.94\linewidth]{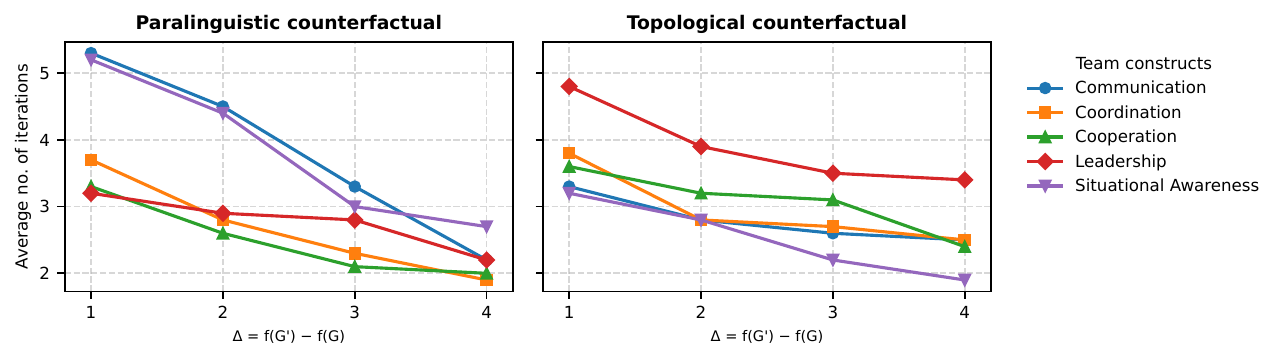}
    \caption{Average number of iterations required to find an appropriate counterfactual using Alg.~\ref{alg:counterfactual}.} 
    \label{fig:counterfactual_plot}
\end{figure*}
Paralinguistic feature extraction relies on accurate speaker diarization. Features are extracted independently for each participant:
non-speaking intervals are masked and assigned neutral values, while eGeMAPS 2.0~\cite{eyben2015geneva} features are extracted for active speakers. To enhance interpretability and enable counterfactual analyses, we selected three features: \emph{average loudness}, which reflects vocal activation and engagement; \emph{harmonics-to-noise ratio}, which captures vocal control and tension; and \emph{alpha ratio}, which is associated with vocal dominance.
Each combination of these three indices defines a behavioral class (Tab.~\ref{tab:behavioral_classes}), allowing each temporal audio window to be mapped to higher-level behavioral abstractions, such as ``Engaged--Cooperative'' or ``Dominant--Aggressive''. Importantly, this discretization is applied only for interpretability; the model retains continuous eGeMAPS values during training to preserve fine-grained acoustic information and predictive fidelity. 
This separation ensures that interpretability does not come at the expense of modeling capacity.  
Textual data from manually corrected transcripts, which include multiple languages, were processed using XLM-RoBERTa~\cite{conneau2020unsupervised}, a multilingual language model pretrained on over 100 languages. Computer vision features were extracted by detecting each team member in three of the five available camera views and encoding their spatial position, movement, and additional nonverbal cues as feature vectors.

\subsection{Baselines}
\label{sec:baselines}

Multiple computational approaches have been proposed to model team behaviors, differing in their ability to capture temporal and relational dependencies in general domains~\cite{Luca2026} and high-stakes domains~\cite{DeLuca2026Actionable}. We organized competitors in terms of the amount of information they are capable of handling (see Modeling Dimension in Tab.~\ref{tab:main_results}). Static Neural Networks (\SNN) model individual features only, and we include them as baselines using minimal information. Temporal Neural Networks (\TNN) capture temporal evolution~\cite{avci2016predicting, gorman2017understanding, yang2020group, 8626131, lin2023interaction}. We considered both recurrent (LSTM~\cite{6795963} and GRU~\cite{cho-etal-2014-learning}) and attention-based (MHA~\cite{NIPS2017_3f5ee243}) strategies. Relational Neural Networks (\ReNN) focus on interactions, modelled via graph neural network architectures~\cite{yang2020group, sharma2023graphitti, li2024dat, malik2023relational}. We considered standard GNN backbones, namely GCN~\cite{kipf2017semi}, GAT~\cite{velickovic2017graph}, and GIN~\cite{xu2018how}. 
Finally, Tempo-Relational Neural Networks (\TReNN)~\cite{Luca2026} combine both temporal and relational aspects, and represent the most expressive class of models. 
We distinguish three alternative strategies to integrate temporal and relational dependencies: (i) \textit{space-then-time}, where relational representations are first computed at each timestep and then processed temporally; (ii) \textit{time-then-space}, where temporal encoding is first applied to node features and subsequently propagated through the graph; and (iii) \textit{joint temporal-relational models}, which jointly capture graph structure and temporal evolution through dynamic graph neural network architectures. 
In particular, for space-then-time and time-then-space \TReNN we varied the spatial modeling paradigm (i.e., GCN, GAT, GIN) and used MHA for temporal modeling, as it outperformed the other strategies in this domain. While in joint tempo-relational models we include TGN, DySAT, and EvolveGCN, which differ in how temporal dynamics are incorporated, ranging from event-based memory mechanisms to temporal attention and evolving graph convolution operators. 
Including these variants enables a systematic comparison across different integration strategies, as summarized in Table~\ref{tab:main_results}.

\begin{figure}
    \centering
    \includegraphics[width=1\linewidth]{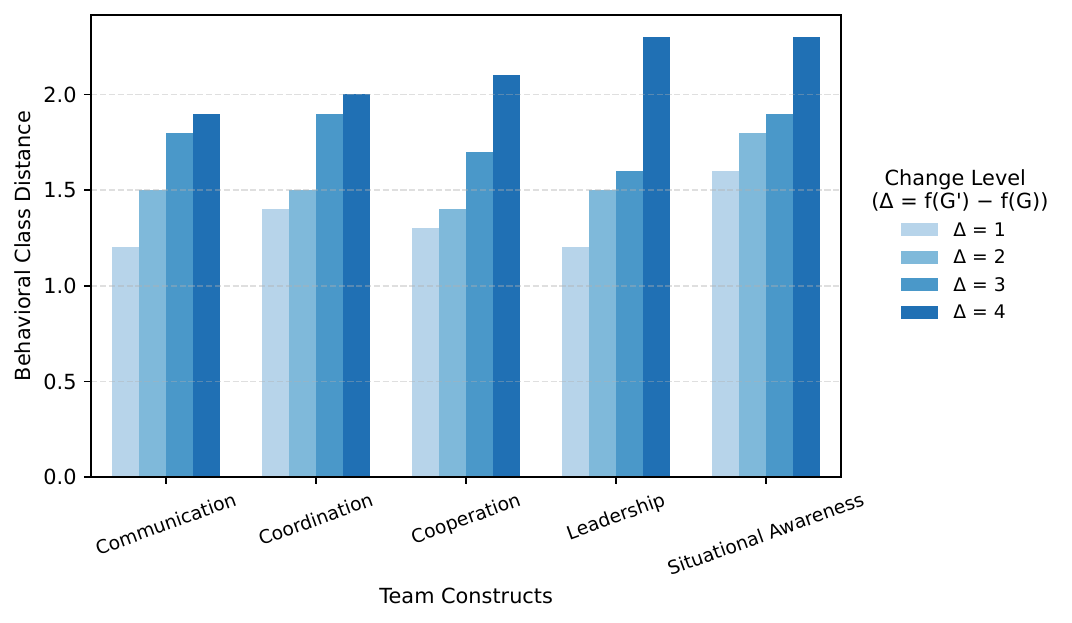}
    \caption{Average distance between the original and counterfactual paralinguistic features of the targeted team member, using Alg.~\ref{alg:counterfactual}.}

    \label{fig:distance_counterfactual}
\end{figure}

\section{Results}

In this section, we first introduce the research questions, followed by the evaluation protocol, implementation details, and experimental results. 
Our work
aims to address the following research questions:
\begin{itemize}
 \item[\textbf{Q1:}] Does \TEReNN outperform existing paradigms in modeling small teams' behavior in a low-data regime?
\item[\textbf{Q2:}] Does \TEReNN support counterfactual algorithms capable of generalizing across topology and features while being interpretable and actionable to optimize teamwork? 
\end{itemize}

\subsection{Predictive performance}

We evaluated the predictive performance of \TEReNN by training it to predict the average annotators' scores for each of the five dimensions of the OTAS behavioral rating schema, and comparing it to the baselines presented in Section~\ref{sec:baselines}. 
We formulated the problem as a multi-class classification task. Models optimize a combination of a cross-entropy and an ordinal loss function, which explicitly accounts for the ordering of the Likert ratings by penalizing predictions proportionally to their distance from the ground-truth class. To mitigate class imbalance, the loss was weighted according to class frequencies, assigning larger penalties to errors on underrepresented classes. The same training objective was adopted for all baselines, and a separate model was trained for each OTAS dimension. We employed a leave-one-team-out evaluation protocol, training on $n-1$ procedures and testing on the single left-out one, and iterating over the left-out procedure. We additionally reserved one training procedure for validation. Table~\ref{tab:main_results} reports the mean and standard deviation of average F1-score for each OTAS dimension, as well as their average, which serves as an overall measure of teamwork.
Results show a substantial advantage of \TEReNN over all baselines across all OTAS dimensions, regardless of the chosen GNN backbone. As expected, the best performance is obtained when \TEReNN is paired with the most expressive backbone (GIN), although differences across backbones remain limited. In conclusion, \TEReNN achieves an improvement of approximately 7\% over the strongest baseline (MHA+GAT). 
The magnitude of this gain, together with its weak dependence on the backbone architecture, highlights the robustness and effectiveness of \TEReNN in the low-data regime. This is particularly evident given that differences among baselines are comparatively modest, even when they leverage varying amounts of information (e.g., \TReNN vs. \TNN or \ReNN).

\begin{figure*}
    \centering
    \includegraphics[width=1\linewidth]{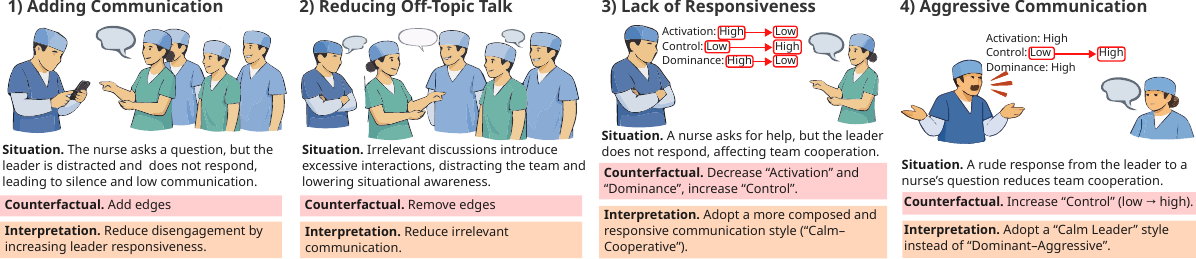}
    \caption{Examples of counterfactuals generated from the dataset. Panels 1) and 2) show topological counterfactuals: in 1), the leader’s distraction leads to poor communication, suggesting a more prompt response; in 2), irrelevant conversation reduces situational awareness, suggesting greater attention and fewer off-topic discussions. Panels 3) and 4) show paralinguistic counterfactuals based on the behavioral classes in Tab.~\ref{tab:behavioral_classes}: in 3), the leader is disengaged from the nurse’s request, reducing team communication, and the counterfactual suggests a calm-cooperative behavior; in 4), the leader exhibits dominant-aggressive behavior toward the nurse, reducing cooperation, and the counterfactual recommends a calm behavioral style.}
    \label{fig:topologicalCongrofactualv2}
    
\end{figure*}

\subsection{Counterfactual assessment}
Understanding how team dynamics can be improved is crucial in high-stakes collaborative settings. One approach is to generate counterfactuals as feedback for the team. The flexibility of \TEReNN makes it particularly suitable for this task by capturing both temporal and relational information: inter-snapshot edges encode temporal dependencies, while intra-snapshot edges capture interactions among team members.
To improve interpretability, we select a subset of high-level features that directly map to observable behaviors (Sec.~\ref{sec:preprocessing}). These include paralinguistic features, whose mapping to semantically meaningful behavioral classes (Table~\ref{tab:behavioral_classes}) enables actionable behavioral suggestions (e.g., remaining calmer or more dominant), and topological features, whose counterfactuals correspond to changes in team interactions.
The counterfactual algorithm proposed in Sec.~\ref{sec:actionable_counterfactual} was evaluated by measuring the iterations required to identify a desired behavioral pattern absent from the original graph. The algorithm searches for the closest exemplar in the embedding space and derives its factual explanation while accounting for team roles and temporal context. 
We grouped counterfactuals by desired score improvement, $\Delta=f(G')-f(G)$, where $f$ is the predictive model, $G$ the original graph, and $G'$ the counterfactual exemplar
By construction, $\Delta>0$, and valid counterfactuals are required to achieve a predicted score of 6 or 7 (Alg.~\ref{alg:counterfactual}).

Counterfactuals were analyzed separately according to the features they modify. For topological changes, we compared communication links between members with the same role in the original and counterfactual graphs. Fig.~\ref{fig:counterfactual_plot} reports the average iterations required to achieve the desired improvement as a function of $\Delta$. Paralinguistic counterfactuals are particularly effective for Leadership and Cooperation, likely because changes in the leader's behavior strongly influence the team, whereas topological counterfactuals perform better for Situational Awareness and Communication, and to a lesser extent Coordination.
Unlike topological counterfactuals, which consist of adding or removing a communication edge, paralinguistic counterfactuals require transitions between behavioral classes (Table~\ref{tab:behavioral_classes}). Fig.~\ref{fig:distance_counterfactual} reports the behavioral class distance between original and counterfactual behaviors. Communication, Coordination, and Cooperation require only limited behavioral changes, even for large score improvements, whereas Leadership and Situational Awareness generally require more substantial modifications, particularly from poor initial performance.

In addition, we conducted a human evaluation with two independent annotators to compare our counterfactual algorithm with the state-of-the-art method CoDy~\cite{qu2025cody}, the strongest competing approach for TReNN. The generated counterfactuals were evaluated in terms of \emph{Realism} (Plausibility), \emph{Punctuality} (Timing), \emph{Helpfulness} (expected performance impact), and \emph{Minimality} (change size). 
All dimensions were rated on a 5-point scale (0 = lowest, 4 = highest).
Our example-based approach achieved higher scores than CoDy in realism (2.3 vs.\ 1.1), punctuality (2.1 vs.\ 1.2), and helpfulness (2.2 vs.\ 0.9), while obtaining a slightly lower score in minimality (2.0 vs.\ 2.3), as CoDy always modifies a single edge. Overall, these results indicate that \TEReNN produces counterfactual suggestions that are perceived as more realistic, timely, and helpful than the current state-of-the-art, with only a minor trade-off in minimality.
\paragraph{Qualitative examples.} The examples in Fig.~\ref{fig:topologicalCongrofactualv2} illustrate the actionable nature of the proposed counterfactual framework across both relational and paralinguistic dimensions. In all four cases, the observed interaction sequence is associated with a low predicted teamwork score, and the counterfactual identifies a minimal, interpretable modification leading to a higher predicted score. In the first case, characterized by poor Communication, the method suggests adding missing communicative links, indicating that the leader should respond more promptly to a nurse’s request. In the second case, associated with low Situational Awareness, the counterfactual recommends removing unnecessary interactions to reduce off-topic communication. The third and fourth examples demonstrate the framework’s ability to operate on paralinguistic behavior. For poor Communication, it recommends shifting toward a more responsive communication style; for poor Cooperation, it suggests reducing an aggressive vocal attitude, moving from a dominant-aggressive to a calmer leadership profile. Overall, these counterfactuals provide concrete, human-interpretable guidance on how team behavior should change, either by modifying interaction patterns or communication style, highlighting the practical value of the proposed approach in high-stakes collaborative settings.
\section{Conclusions}
In this work, we introduced \TEReNN, a Time-Expanded Relational Neural Network for multimodal modeling of small surgical teams in low-data settings. By embedding temporal evolution directly into the graph structure, the proposed approach jointly captures interpersonal interactions and their dynamics over time, leading to consistent improvements over strong static, temporal, relational, and tempo-relational baselines across all OTAS teamwork dimensions.
Building on this representation, we also proposed an exemplar-based counterfactual approach for actionable reasoning. Unlike post-hoc explanations that only justify model predictions, our method identifies minimal and interpretable behavioral changes grounded in real observations, providing actionable suggestions at both relational and paralinguistic levels. The qualitative examples further show that these counterfactuals can be translated into concrete recommendations to improve teamwork dimensions in high-stakes team settings.
Overall, our results demonstrate that Time-Expanded multimodal graph representations constitute a promising foundation for actionable AI in collaborative environments.

\begin{acks}
Funded by the European Union. Views and opinions expressed are however those of the author(s) only and do not necessarily reflect those of the European Union or the European Health and Digital Executive Agency (HaDEA). Neither the European Union nor the granting authority can be held responsible for them. Grant Agreement no. 101120763 - TANGO. GV and VMDL acknowledge the support of the MUR PNRR project FAIR - Future AI Research (PE00000013) funded by the NextGenerationEU. AL was supported by the Research Council of Norway through its Centre of Excellence Integreat - The Norwegian Centre for knowledge-driven machine learning, project number 332645. We acknowledge Raffaella Sabrina Fellone and Vincenza Moncelli for supporting the annotation process.
\end{acks}

\bibliographystyle{ACM-Reference-Format}
\balance
\bibliography{sample-base}
\end{document}